\documentclass[letterpaper,10pt,conference]{ieeeconf}

\IEEEoverridecommandlockouts
\usepackage{amsmath,amssymb,mathtools}
\usepackage{booktabs}
\usepackage{balance}
\usepackage{amsmath,amssymb}
\usepackage{array,booktabs,tabularx}
\usepackage{graphicx}
\usepackage{url}
\usepackage{makecell}
\newcommand{\R}{\mathbb{R}}
\newcommand{\SE}{\mathrm{SE}}

\newcommand{\cQ}{\mathcal{Q}}

\newcommand{\cO}{\mathcal{O}}
\newcommand{\cF}{\mathcal{F}}

\newcommand{\cT}{\mathcal{T}}
\newcommand{\cX}{\mathcal{X}}

\title{From Language to Task Maps: Compiling Semantic Relations While Preserving Task-Relevant Freedom}

\author{Jaegyun Park$^{1,*}$, Jingwang Lee$^{1,*}$, Jungsoo Lee$^{1}$, Soonwoong Hwang$^{2}$ and Wansoo Kim$^{2,\dagger}$%
\thanks{$^{*}$These authors contributed equally to this work.}%
\thanks{$^{\dagger}$Corresponding author: wansookim@hanyang.ac.kr}%
\thanks{$^{1}$Department of Robotics, Hanyang University, Seoul, Republic of Korea.
        \{chant29, jingwang1203, lpigeon\}@hanyang.ac.kr}%
\thanks{$^{2}$Department of Robotics,
        Hanyang University ERICA, Ansan, Republic of Korea.}%
}

\begin{document}
\maketitle

\begin{abstract}
    Natural-language manipulation instructions specify qualitative relations, whereas continuous controllers require state-evaluable task quantities, differentials, and completion conditions. Because a qualitative relation generally leaves part of the relative configuration unspecified, expanding it into a complete pose can introduce unintended constraints. We present a typed semantic-to-geometric interface in which language specifies entities, relations, and phases, while each relation indexes a registered specification of its task-relevant distinctions and preserved freedoms. A robot-side compiler grounds these specifications, constructs relation-specific task maps and consistent differentials using conformal geometric algebra, and composes the resulting policies through RMPflow. To evaluate the division of responsibility between the language model and the compiler, we compared a Semantic Topology interface with one that additionally requires relation-specific geometric specifications over 60 instructions. Both produced correct shared semantic content in 41/60 cases, but critical errors under their respective interface requirements occurred in 19/60 and 58/60 cases. Across 64 grounded evaluations spanning eight geometric relation forms, the task maps preserved registered null directions and responded to relation-relevant perturbations; analytic directional derivatives agreed with finite differences, and Jacobian ranks matched the registered dimensions. In three closed-loop ablations using a simulated Franka Emika Panda in MuJoCo, fixing a relation-preserved coordinate increased median terminal progress error by 20.24--71.00~mm while the retained relation errors remained within their evaluation bounds. These results support compiling relation-visible geometry and preserved freedom together into composable continuous objectives.
\end{abstract}

\section{Introduction}
\label{sec:introduction}

Large language models (LLMs) have been used to interpret natural-language manipulation instructions, select executable skills, and generate long-horizon task plans or robot programs~\cite{ichter2023saycan,driess2023palme,singh2023progprompt,liang2023code}.
However, the representations in which language expresses manipulation objectives differ from those required by continuous control. Natural-language commands typically specify qualitative relations among entities, such as aligning a tool with a surface, keeping an object upright, or moving along a path. A controller, in contrast, requires task quantities, their differentials, and completion conditions that can be evaluated from the current robot configuration and the geometric state of the scene. The representation bridging this gap therefore determines which geometric distinctions are exposed to the controller. This work addresses the following question: \emph{How can a qualitative, relational manipulation specification be converted into composable continuous control objectives while preserving the geometric distinctions and freedoms implied by each relation?}

This translation cannot be reduced to expanding every relation into a complete relative pose, because a manipulation relation generally leaves part of the relative configuration unspecified. Even when the same tool axis is represented as a Line, parallelism observes only a directional discrepancy and does not distinguish spatial offset, whereas coincidence must distinguish both direction and transverse offset. Replacing such relation-specific objectives with full-pose targets constrains freedoms that the task did not specify and can create representation-induced conflicts between otherwise compatible objectives. Which quantities are observed and which freedoms are preserved must therefore be determined by the relation rather than by the primitive type alone: states that a relation does not distinguish should produce the same task quantity in control.

\begin{figure*}[t]
    \centering
    \includegraphics[width=\linewidth]{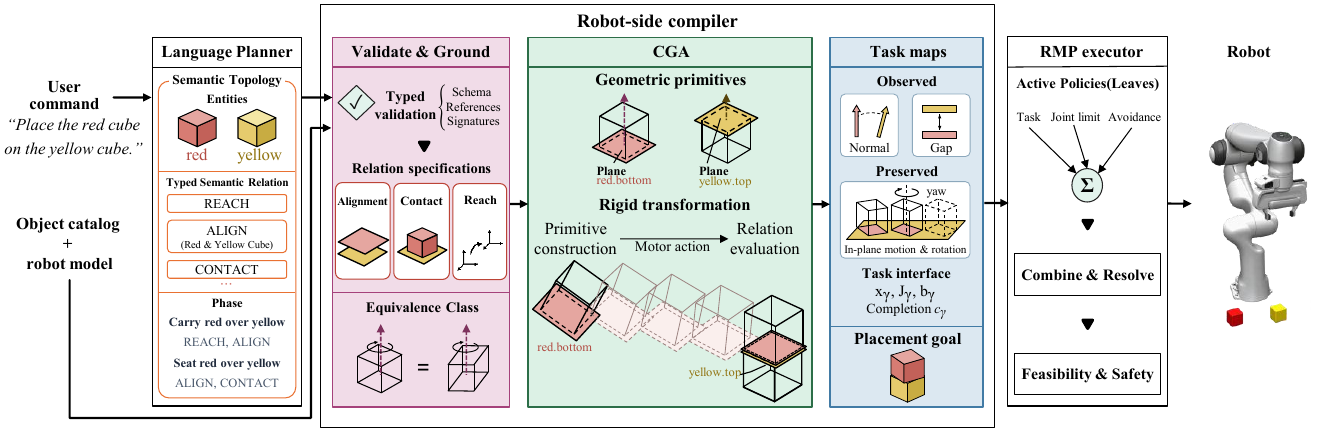}
    \caption{Overview of the semantic-to-geometric compilation pipeline. The language planner specifies entities, relations, and phases; the robot-side compiler selects and grounds relation-visible geometry before the resulting task maps are composed for continuous control.}
    \label{fig:ICRA_framework_figure}
    \vspace{-0.3cm}
\end{figure*}

We introduce a Semantic Topology---a typed task graph specifying participating entities, required relations, and their activation across ordered phases---as the interface between language and continuous control. Each typed semantic relation $\gamma$ indexes a registered specification $\Gamma_\gamma$ that defines its task-relevant distinctions, preserved freedoms, and completion predicate. A robot-side compiler binds this specification to an object catalog, a calibrated robot model, and the runtime observation to produce a relation-specific task map and its differential. Construction and transformation of heterogeneous primitives use conformal geometric algebra (CGA) as a common carrier~\cite{dorst2007geometric,low2023geometric,low2025gafro}, and the relation-local policies are composed in configuration space through RMPflow~\cite{ratliff2018rmp,cheng2021rmpflow}. Objectives such as directional alignment, contact, path traversal, and relative-pose maintenance can then be composed while each retains its own freedoms. For example, line coincidence leaves axial translation available to an insertion policy. Figure~\ref{fig:ICRA_framework_figure} summarizes the overall semantic-to-geometric compilation pipeline.

The contributions of this work are:
\begin{enumerate}
    \item \textbf{A typed semantic relation specification that makes task-relevant distinctions and preserved freedoms explicit.} 
    Language specifies entities, typed semantic relations, and phases, while each registered relation defines which geometric distinctions are visible to control and which freedoms remain unspecified. We formalize this semantics through relation-induced equivalence, such that configurations not distinguished by a relation produce the same task quantity.

    \item \textbf{A factorized compiler that realizes relation semantics without reintroducing preserved freedoms as constraints.}
    Using shared CGA primitive operations, the compiler constructs relation-specific task maps and consistent differentials whose null directions remain unweighted by that relation after RMPflow pullback.
\end{enumerate}

\section{Related Work}
\label{sec:related_work}

\subsection{Language-Guided Manipulation and Geometric Constraint Generation}

Representative language-guided robotics systems use language models as grounded skill selectors or task-level planners. SayCan combines language-model scores with affordance values to select executable skills~\cite{ichter2023saycan}, whereas PaLM-E jointly processes continuous sensor observations and language embeddings in an embodied multimodal model~\cite{driess2023palme}. ProgPrompt and Code as Policies generate context-dependent task plans or executable programs~\cite{singh2023progprompt,liang2023code}. These works primarily address language-level planning and skill composition rather than treating the geometric distinctions exposed by a selected manipulation relation as an explicit interface specification.

Recent work has extended this direction by generating spatial objectives from language and vision. VoxPoser uses LLMs and vision-language models to construct composable 3-D value maps and plans end-effector trajectories with a model-based planner~\cite{huang2023voxposer}. ReKep represents relational constraints as numerical Python cost functions over semantic keypoints and solves them through hierarchical optimization~\cite{huang2024rekep}. GeoManip derives stage-specific geometric constraints from object-part relations for trajectory optimization~\cite{tang2025geomanip}. RelAfford6D grounds relational affordance graphs into metric $\SE(3)$ poses and kinematic constraint manifolds~\cite{zhang2026relafford6d}. These methods demonstrate effective ways to generate geometric objectives from language and perception. Our question is narrower: after a semantic relation has been selected, should the language model explicitly author its observed geometric components and preserved freedoms for every instruction, or should the compiler retrieve them from a registered relation specification? E1 in Sec.~\ref{sec:experiments} compares these two interfaces directly.

Task and Motion Planning (TAMP) integrates discrete task decisions with continuous geometric feasibility~\cite{garrett2021tamp}, while Logic-Geometric Programming combines symbolic skeletons with geometric optimization~\cite{toussaint2015lgp}. PRoC3S casts the open parameters and environmental constraints of LLM-generated, continuously parameterized skill programs as a continuous constraint-satisfaction problem~\cite{curtis2025proc3s}. These approaches solve continuous feasibility and open parameters under a given symbolic structure or skill program. Our work is complementary and addresses the preceding step of representation selection: determining which geometric quantities a selected relation exposes to optimization or control.

\subsection{Constraint-Based Task Representations and Task-Relevant Freedom}

Manipulation planning has long represented objectives as sets of admissible configurations or constraint manifolds rather than complete poses. Constraint-manifold planning searches for motion on lower-dimensional feasible sets~\cite{berenson2009constraint}, and Task Space Regions represent a goal as an admissible pose region to preserve task freedom~\cite{berenson2011tsr}. Task-manifold learning represents constrained object manipulation through learned manifolds~\cite{englert2018taskmanifold}, while object-symmetry-aware pose representations treat poses describing the same physical configuration as equivalent~\cite{bregier2018pose}. Sequence-of-Constraints model predictive control (MPC) formulates sequential manipulation through constraint activation and reactive control~\cite{toussaint2022socmpc}.

We connect these views of admissible freedom and geometric equivalence to language-authored typed semantic relations. Unlike object symmetry, which is defined by object identity, the equivalence considered here is induced by the active relation (Sec.~\ref{sec:equivalence}). A registered relation specification therefore fixes relation-induced equivalence together with its task quantity, completion predicate, and differential.

\subsection{Geometric Representations and Compositional Motion Policies}

CGA represents Points, Lines, Planes, Spheres, Circles, and rigid transformations in a common algebraic framework~\cite{dorst2007geometric}. It has also been used to formulate geometric objectives, Jacobians, and optimal-control quantities for manipulation~\cite{low2023geometric}. The \texttt{gafro} framework provides reusable geometric-algebra computation for robotics~\cite{low2025gafro}. We use CGA as a common carrier for primitive construction and transformation, whereas the registered geometric relation form decides which residual, margin, or progress coordinate is exported from a CGA expression.

Riemannian Motion Policies combine task-space motion objectives with state-dependent metrics~\cite{ratliff2018rmp}, and Riemannian Motion Policy flow (RMPflow) composes multiple local policies in configuration space through task-map differentials and geometric pullback~\cite{cheng2021rmpflow}. Geometric Fabrics likewise use geometric structure to synthesize multiple behaviors~\cite{vanwyk2022fabrics}. These methods compose task maps and local policies after they have been defined. Our compiler operates upstream, converting language-authored relations into those task maps and differentials.

\section{Methodology}
\label{sec:methodology}

Our objective is to convert natural-language manipulation instructions
into continuous control while preserving the freedoms left unspecified
by each typed semantic relation. A language planner first constructs a
Semantic Topology of entities, relations, and phases, after which the
robot-side compiler grounds each relation into geometric primitives,
relation-specific CGA task quantities, consistent differentials, and
continuous policies.

\subsection{Semantic Topology and Relation Retrieval}
\label{sec:topology}

Let $u$ denote a language instruction, $\mathcal W^{\mathrm{sem}}$ a semantic world description, $q\in\cQ$ the robot configuration, and $o$ a runtime observation. Given $u$, the LLM receives $\mathcal W^{\mathrm{sem}}$ listing available entities and features, the registered relation vocabulary (Table~\ref{tab:semantic-vocabulary}), and an output schema. Its output is the Semantic Topology $G_s$, a JSON task specification that selects relation types from the vocabulary, binds them to entities, supplies mode arguments, and assigns their activation to ordered phases:
\begin{equation}
\begin{aligned}
G_s&=(E,R,S),
&R&=R_{\mathrm{cont}}\mathbin{\dot\cup}R_{\mathrm{event}},\\
S&=(A_1,\ldots,A_{N_p}),
&\gamma&=(\tau,e_s,f_s,e_r,f_r,a_\gamma,\mathcal U),
\end{aligned}
\label{eq:semantic-topology}
\end{equation}
\noindent where $E$ identifies task entities such as objects, parts, robot links, and reference frames, $R_{\mathrm{cont}}$ contains continuous relations to be achieved or maintained, whereas $R_{\mathrm{event}}$ contains discrete directives, such as grasp and release, that change the interaction mode. $S$ is an ordered sequence of phases, and $A_k\subseteq R$ is the set of relations active simultaneously in phase $k$. Each selected relation is encoded as a typed semantic relation $\gamma$, containing its registered type $\tau$, subject and reference entity--feature bindings, mode arguments $a_\gamma$ such as the alignment mode or gripper command, and user-specified constraints $\mathcal U$. $\mathcal U$ carries only distances or angles the user actually stated; no other numerical value is placed in the topology.

\begin{table}[t]
\caption{Typed semantic relations and their meanings.}
\label{tab:semantic-vocabulary}
\centering
\footnotesize
\setlength{\tabcolsep}{4pt}
\renewcommand{\arraystretch}{1.08}
\newcommand{\rowgap}{\addlinespace[2pt]}
\begin{tabularx}{\columnwidth}{@{}>{\raggedright\arraybackslash}p{0.44\columnwidth}>{\raggedright\arraybackslash}X@{}}
\toprule
Typed semantic relation & Meaning \\
\midrule
\texttt{REACH\_\allowbreak AFFORDANCE} (RA)
& Bring the subject to a reference affordance \\
\rowgap
\texttt{ALIGN\_\allowbreak AFFORDANCES} (ALN)
& Align paired features under the selected condition \\
\rowgap
\texttt{MAINTAIN\_\allowbreak DIRECTION} (MD)
& Keep a body axis aligned with a reference direction \\
\rowgap
\texttt{ESTABLISH\_\allowbreak CONTACT} (EC)
& Bring the subject into contact with its support \\
\rowgap
\texttt{MAINTAIN\_\allowbreak AFFORDANCE\_\allowbreak CONTACT} (MAC)
& Maintain a grasp or designated surface contact \\
\rowgap
\texttt{CLEAR\_\allowbreak SUPPORT} (CS)
& Move the subject clear of its support \\
\rowgap
\texttt{MAINTAIN\_\allowbreak RELATIVE\_\allowbreak POSE} (MRP)
& Preserve the relative pose captured at activation \\
\rowgap
\texttt{TRAVERSE\_\allowbreak AFFORDANCE} (TA)
& Move a subject feature along a reference path \\
\rowgap
\texttt{ACTUATE\_\allowbreak AFFORDANCE} (ACT)
& Drive an articulation to a named target state \\
\rowgap
\texttt{AVOID\_\allowbreak COLLISION} (AC)
& Avoid contacts forbidden by the contact policy \\
\rowgap
\texttt{GRIPPER\_\allowbreak COMMAND}
& Open or close the gripper (discrete directive) \\
\rowgap
\bottomrule
\end{tabularx}
% \vspace{-0.2cm}
\end{table}

\begin{table*}[t]
\caption{Representative geometric relation forms: task quantities, preserved freedoms, and expected ranks $\dim\mathcal B_\gamma$.}
\label{tab:relation-forms}
\centering
\footnotesize
\renewcommand{\arraystretch}{1.08}
\begin{tabularx}{\textwidth}{@{}p{0.16\textwidth}p{0.15\textwidth}p{0.26\textwidth}Xc@{}}
\toprule
Geometric relation form & Primitive inputs & Task quantity & Preserved freedom & $\dim\mathcal B_\gamma$ \\
\midrule
Line parallelism
& Line--Line
& Direction residual
& All translations and rotation about the Line axis & 2 \\
\addlinespace[2pt]
Line coincidence
& Line--Line
& Direction and transverse-offset residual
& Translation along the common axis and axial spin & 4 \\
\addlinespace[2pt]
Plane-normal alignment
& Plane--Plane
& Normal-alignment residual
& Normal gap and in-plane pose & 2 \\
\addlinespace[2pt]
Plane coincidence
& Plane--Plane, $\mathcal U$
& Normal and signed-gap residual (target gap $d^{*}\in\mathcal U$, default 0)
& In-plane pose & 3 \\
\addlinespace[2pt]
Directional alignment
& Line--reference direction
& Axis-direction residual
& Translation and rotation about the aligned axis & 2 \\
\addlinespace[2pt]
Signed gap or contact
& Point--Plane
& Signed normal distance or margin
& Tangential directions of the Plane & 1 \\
\addlinespace[2pt]
Circle incidence
& Point--Circle
& Off-curve incidence residual
& Tangential motion along the Circle & 2 \\
\addlinespace[2pt]
Circle traversal
& Point--Circle, $z_\gamma$
& Incidence and path progress
& Orientation not otherwise specified & 3 \\
\bottomrule
\end{tabularx}
\vspace{-0.4cm}
\end{table*}

Before grounding, typed validation checks schema consistency, entity and phase references, relation signatures, required fields, and declared capabilities; invalid specifications are rejected with diagnostics. For each continuous relation in $R_{\mathrm{cont}}$, the compiler then uses its type, bound feature metadata, and mode arguments to retrieve a registered specification $\Gamma_\gamma=(\cO_\gamma,\mathcal N_\gamma,I_\gamma,\dim\mathcal B_\gamma,c_\gamma)$: the relation operation, normalization, and coefficient index that define the task quantity (Sec.~\ref{sec:equivalence}), the registered rank of the relation-visible representation, and the completion predicate used at phase transitions (Sec.~\ref{sec:composition}). Together these fix the geometric relation form; Table~\ref{tab:relation-forms} lists representative forms. The compiler grounds the primitive operands using the catalog $\mathcal K$, the calibrated robot model $\cF_r$, and the configuration and observation $q_0,o_0$ at activation:
\begin{equation}
\gamma
\xmapsto{\;C_{\mathrm{type}}\;}\Gamma_\gamma
\xmapsto[\mathcal K,\cF_r,q_0,o_0]
{\;C_{\mathrm{ground}}\;}\cT_\gamma
=\bigl(\Gamma_\gamma,X_s^0,X_r^0,z_\gamma\bigr).
\label{eq:grounding}
\end{equation}
$C_{\mathrm{type}}$ selects $\Gamma_\gamma$ without producing numerical values. $C_{\mathrm{ground}}$ binds that specification to the current scene, producing the local subject and reference primitives $X_s^0,X_r^0$ and a discrete branch state $z_\gamma$. The branch state holds discrete choices that stay fixed over one smooth execution interval, such as the alignment sign of a sign-invariant axis or the progress branch along a path. An unregistered signature or an invalid operand produces a diagnostic rather than an arbitrary target pose. User-specified distances and angles remain in $\mathcal U$, whereas embodiment- and scene-dependent quantities, including grasp poses, feasible inverse-kinematics branches, collision clearances, and controller metrics, are resolved during grounding and execution.

In the stacking example of Fig.~\ref{fig:ICRA_framework_figure}, the planner binds \texttt{ALN} to \texttt{red.bottom} and \texttt{yellow.top} with alignment mode plane coincidence. Since both features are Planes, $C_{\mathrm{type}}$ retrieves the Plane-coincidence form of Table~\ref{tab:relation-forms}: a normal and signed-gap residual with default $d^{*}=0$, in-plane pose preserved, and $\dim\mathcal B_\gamma=3$. $C_{\mathrm{ground}}$ instantiates $X_s^0$ from the bottom Plane of the red block and $X_r^0$ from the top Plane of the yellow block in $\mathcal K$. The carry phase activates \texttt{RA} and \texttt{ALN}, whereas seating activates \texttt{ALN} and \texttt{EC}, whose Point--Plane form adds a contact margin without constraining the in-plane pose that \texttt{ALN} leaves free.

\subsection{Relation-Induced Equivalence and CGA Realization}
\label{sec:equivalence}

Let the grounded primitive state be $\xi_\gamma=F_\gamma(q,o;z_\gamma)\in\cX_\gamma$, and write $\xi\sim_\gamma\xi'$ when the relation does not distinguish two states. The relation-visible state space and task representation satisfy
\begin{equation}
\begin{aligned}
\mathcal B_\gamma
&=\cX_\gamma/\!\sim_\gamma,
&\pi_\gamma&:\cX_\gamma\rightarrow\mathcal B_\gamma,\\
\phi_\gamma
&=\bar\phi_\gamma\circ\pi_\gamma,
&\xi\sim_\gamma\xi'
&\Longrightarrow\phi_\gamma(\xi)=\phi_\gamma(\xi').
\end{aligned}
\label{eq:quotient}
\end{equation}
The robot-side task map composes grounding with this representation,
\begin{equation}
\psi_\gamma=\bar\phi_\gamma\circ\pi_\gamma\circ F_\gamma:\;
(q,o)\xrightarrow{F_\gamma}\cX_\gamma\xrightarrow{\pi_\gamma}\mathcal B_\gamma\xrightarrow{\bar\phi_\gamma}\R^{m_\gamma}.
\label{eq:task-map-chain}
\end{equation}
Equation~\eqref{eq:quotient} states two separate things. First, states that the relation treats as equivalent produce the same task quantity; this invariance is guaranteed by constructing $\phi_\gamma$ to factor through $\pi_\gamma$. Second, the representation should not lose distinctions the relation makes. Globally this would require $\bar\phi_\gamma$ to be injective over the admissible domain; we claim only the differential condition on each regular branch,
\begin{equation}
\ker D\phi_\gamma(\xi)=\ker D\pi_\gamma(\xi),
\label{eq:quotient-kernel}
\end{equation}
which holds when $\bar\phi_\gamma$ is an immersion on that branch, in which case $\operatorname{rank}D\phi_\gamma=\dim\mathcal B_\gamma$. Registering $\dim\mathcal B_\gamma$ with each geometric relation form, together with the null-direction tests of E2, therefore makes condition~\eqref{eq:quotient-kernel} checkable: the implemented Jacobian must introduce no null direction beyond $\ker D\pi_\gamma$, so its rank must equal the registered value. E2 in Sec.~\ref{sec:experiments} evaluates this condition numerically. In the stacking example, $\xi_\gamma$ consists of the two bound Planes. Under Plane-normal alignment, $\xi\sim_\gamma\xi'$ whenever the states differ by any translation or by rotation about the normal, so $\mathcal B_\gamma$ retains only two direction coordinates. Under Plane coincidence on the same Planes, translation along the normal is no longer equivalent and $\dim\mathcal B_\gamma=3$, leaving the in-plane pose. The equivalence is thus fixed by the geometric relation form rather than by the objects, in contrast to object symmetry~\cite{bregier2018pose}: the same blocks induce different quotients under the two forms.

CGA is a computational carrier that constructs and transforms Points, Lines, Planes, Circles, and Frames through common operations. A local primitive $X_a^0$ is transformed at runtime by a motor $M_a$,
\begin{equation}
X_a(q,o)=M_a(q,o)X_a^0\widetilde M_a(q,o),\qquad a\in\{s,r\}.
\label{eq:cga-transform}
\end{equation}
The registered relation operation is applied to the normalized subject and reference operands, and only the task-visible coefficients are extracted:
\begin{equation}
x_\gamma
=\psi_\gamma(q,o;z_\gamma)
=\operatorname{coeff}_{I_\gamma}
\Bigl(\mathcal N_\gamma\bigl(\cO_\gamma(\widehat X_s,\widehat X_r;z_\gamma)\bigr)\Bigr).
\label{eq:relation-evaluation}
\end{equation}
The operation $\cO_\gamma$, normalization $\mathcal N_\gamma$, and coefficient index $I_\gamma$ come from $\Gamma_\gamma$; CGA supplies only the shared transformation~\eqref{eq:cga-transform} and coefficient extraction, whereas $\Gamma_\gamma$ determines which motions leave $x_\gamma$ unchanged.

\subsection{Differential Task Interface and Freedom Preservation}
\label{sec:differential}

% [vs main.tex] Moving-reference terms kept as in the original (tracked dynamic obstacles and moving targets appear in the executions).
For each smooth execution interval, we hold the grounded parameters and $z_\gamma$ fixed and assume that $\psi_\gamma$ is $C^2$ over the operating domain. Differentiating the complete exported task map gives the velocity and acceleration interfaces, including a moving reference:
\begin{equation}
\begin{gathered}
J_\gamma=D_q\psi_\gamma(q,o;z_\gamma),\\
\dot x_\gamma=J_\gamma\dot q+\partial_t\psi_\gamma,
\qquad
\ddot x_\gamma=J_\gamma\ddot q+b_\gamma,\\
b_\gamma=D_q^2\psi_\gamma[\dot q,\dot q]
+2D_q(\partial_t\psi_\gamma)\dot q
+\partial_{tt}\psi_\gamma.
\end{gathered}
\label{eq:differential-interface}
\end{equation}
Here, $\partial_t$ is a partial derivative with $q$ held fixed. For a time-independent task, $b_\gamma=\dot J_\gamma\dot q$. The derivative $J_\gamma$ is obtained by differentiating through the primitive transformation, the CGA relation operation, normalization, and coefficient selection. We therefore compute the residual and Jacobian from the same task map, avoiding a separate Euclidean approximation that could describe a different task. E2 checks this consistency by comparing the analytic $J_\gamma$ against finite differences.

Let $v\in T_q\cQ$ be a continuous variation that the relation does not distinguish, and let $M_\gamma\succeq0$ be the metric of the relation leaf. Factorization and the chain rule give the condition
\begin{equation}
D\pi_\gamma\,D_qF_\gamma[v]=0
\;\Longrightarrow\;
J_\gamma v=0
\;\Longrightarrow\;
v^\top J_\gamma^\top M_\gamma J_\gamma v=0.
\label{eq:semantic-null}
\end{equation}
When the implemented residual and Jacobian preserve this factorization, a relation-irrelevant variation remains in the task-map kernel and receives no metric weight after RMPflow pullback, leaving the freedom available to other policies. E2 evaluates this condition through null-direction responses and differential consistency.

\subsection{RMPflow Composition and Phase Execution}
\label{sec:composition}

Each active geometric relation produces a Riemannian Motion Policy in natural form $(f_i,M_i)$, where $M_i$ is a positive-semidefinite metric and $f_i$ is the metric-weighted desired acceleration~\cite{ratliff2018rmp}. During phase $k$, the grounded relation leaves and the robot-local leaves are collected in $\mathcal L_k=(A_k\cap R_{\mathrm{cont}})\cup\mathcal L_{\mathrm{robot}}$, where $\mathcal L_{\mathrm{robot}}$ contains posture and collision-avoidance leaves. The leaves are composed in configuration space through RMPflow~\cite{cheng2021rmpflow}:
\begin{equation}
\begin{gathered}
M_q=\sum_{i\in\mathcal L_k}J_i^\top M_iJ_i,\\
f_q=\sum_{i\in\mathcal L_k}
J_i^\top\bigl(f_i-M_ib_i\bigr),\\
\ddot q_{\mathrm{nom}}=M_q^\dagger f_q.
\end{gathered}
\label{eq:rmp-composition}
\end{equation}
The nominal acceleration $\ddot q_{\mathrm{nom}}$ is the output of relation composition; the final actuator command is generated after robot-side feasibility constraints are applied. Guard-enabled discrete directives such as grasp and release are handled by a separate event executor.

Phase transitions jointly evaluate equality tolerances, active inequality margins, progress thresholds, and required event acknowledgments. Let $C_k^{\mathrm{geom}}:=\bigwedge_{\gamma\in A_k\cap R_{\mathrm{cont}}}c_\gamma(x_\gamma,o)$ denote satisfaction of all active geometric conditions and $C_k^{\mathrm{event}}:=\bigwedge_{\rho\in A_k\cap R_{\mathrm{event}}}c_\rho^{\mathrm{event}}(o)$ acknowledgment of all active events. Then
\begin{equation}
k^+=k+1
\quad\Longleftrightarrow\quad
C_k^{\mathrm{geom}}\land C_k^{\mathrm{event}},
\label{eq:phase-transition}
\end{equation}
and if this guard is not satisfied the active phase remains $k$.

\subsection{Scope of Guarantees}
\label{sec:scope}

The compiler provides only local, per-relation guarantees. Each relation assigns the same task quantity to equivalent states by construction, observes at the differential level only the directions the relation distinguishes on a regular branch, and assigns no metric weight to directions it does not observe. What the compiler does not guarantee is the global feasibility of mutually conflicting relations, convergence of a phase along an executed trajectory, or closed-loop stability. Safety constraints are applied in a separate layer after relation composition and lie outside the claims of this section.

\section{Experimental Evaluation}
\label{sec:experiments}

\subsection{Experimental Setup and Evidence Protocol}

E1 evaluates the division of responsibility between language and the compiler. E2 tests relation-map consistency, cross-task reuse, and whether constraining a preserved freedom obstructs concurrent progress. E3 examines prompt-dependent relation selection and execution in Nut insertion.

Closed-loop experiments use a Franka Emika Panda in MuJoCo, and task instructions are issued through ROS-MCP~\cite{rosmcp}. The language-facing interface exposes the registered entities, features, relations, and Semantic Topology schema. The resulting entity--relation--phase graphs are executed through the same robot-side compiler and RMP controller.

\subsection{E1: Relation-Interface Responsibility}

\subsubsection{Design}

E1 compares two interfaces under the same instructions, public scene descriptions, registered affordances, model settings, and isolated-session protocol. All 120 calls use \texttt{gpt-5.6-sol} with a 180~s response-time limit; no explicit sampling seed or temperature is supplied. Under the relation-level condition $\mathrm{R}$, the language model specifies entities, semantic relations, user-authored constraints, and phase activation, while delegating relation-visible geometry to the compiler. Under the geometry-explicit condition $\mathrm{G}$, the model must additionally specify the observed and constrained components, preserved freedoms, condition kind, and activation scope for every relation. Selecting a registered condition label belongs to the shared semantic content; explicitly describing its geometric components and freedoms is the additional requirement under $\mathrm{G}$.

The evaluation suite comprises 11 fixed semantic categories: Line parallelism and coincidence; Plane-normal alignment and signed offset; directional alignment and relative-pose orientation; Line and Circle traversal; and three multi-relation compositions. It contains ten paraphrases for Line parallelism and five for each remaining category, giving 60 paired instruction items and 120 isolated language-model calls. Prompts, schemas, public scene descriptions, and a compiler-independent oracle are fixed before evaluation. The oracle specifies the required shared semantic content for both conditions and the required and prohibited geometric commitments for $\mathrm{G}$. A deterministic scorer compares relation content rather than author-chosen relation or phase identifiers, treats the oracle entity fields as required subsets, and retains parameter and activation-order mismatches as errors. Every scheduled call remains in the denominator, including a timeout or invalid response.

The primary outcome is a critical interface error. For $\mathrm{R}$, this denotes an error in the shared semantic content. For $\mathrm{G}$, it additionally includes a missing or contradictory geometric specification, an unsupported commitment, a relation-irrelevant constraint, or a missing response. Under $\mathrm{R}$, geometry delegated to the compiler is not scored as if it had been explicitly and correctly authored by the language model.

\subsubsection{Results}

\begin{table}[!htbp]
\caption{Paired relation-interface results ($N=60$ per condition);\\ --- denotes a quantity delegated to the compiler.}
\label{tab:e1-results}
\centering
\footnotesize
\renewcommand{\arraystretch}{1.1}
\begin{tabular}{@{}lrr@{}}
\toprule
Outcome & $\mathrm{R}$ & $\mathrm{G}$ \\
\midrule
Schema-valid response within time & 60/60 & 57/60 \\
\addlinespace[2pt]
Exact shared semantic content & 41/60 & 41/60 \\
\addlinespace[2pt]
Exact geometric specification & --- & 2/60 \\
\addlinespace[2pt]
Critical interface error & 19/60 & 58/60 \\
\bottomrule
\end{tabular}
\vspace{-0.1cm}
\end{table}

Table~\ref{tab:e1-results} summarizes the outcomes; the three missing responses under $\mathrm{G}$ were timeouts. The equal shared-semantic count (41/60) does not imply that the same items were correct in both conditions. Under the full interface requirements, critical errors occurred in 19/60 cases under $\mathrm{R}$ and 58/60 under $\mathrm{G}$. The primary-error breakdown was 19 shared-semantic errors under $\mathrm{R}$, and 16 shared-semantic errors, 26 geometric-specification errors, 13 prohibited geometric commitments, and three timeouts under $\mathrm{G}$.

Because both conditions achieved the same shared semantic-content accuracy, this result does not indicate a difference in the model's ability to select relations. Instead, it shows that requiring the language model to restate an already registered geometric specification for every call increases the opportunities for interface error. E1 therefore supports assigning relation-specific geometry to the registered compiler, but it does not show that language models cannot reason about geometry, nor does it compare closed-loop task success. Since semantic errors remained in 19/60 relation-level outputs, generated topologies are not executed directly but pass through the validator of Sec.~\ref{sec:topology}.

\subsection{E2: Relation Geometry and Composition}

E2 tests whether the implemented relation maps preserve their registered task-visible distinctions and freedoms through local evaluation, differentiation, and composition.

\subsubsection{Task-Map Invariance and Differential Consistency}

We evaluate all eight geometric relation forms in Table~\ref{tab:relation-forms} at eight grounded configurations each, sampled from task executions, giving 64 evaluations. At each configuration, a relation-null perturbation should leave the exported task quantity unchanged, whereas a relation-relevant perturbation should remain observable. For a continuous null direction $v_{\sim}$, we test the condition of Sec.~\ref{sec:differential}:
\begin{equation}
J_\gamma v_{\sim}\approx0,\qquad
v_{\sim}^{\top}J_\gamma^{\top}M_\gamma J_\gamma v_{\sim}\approx0.
\label{eq:e2-null}
\end{equation}
Here, $J_\gamma$ is computed with respect to local displacement coordinates of the subject primitive rather than $q$, allowing its rank to be compared directly with $\dim\mathcal B_\gamma$. Finite differences check the selected null and relevant directions. Probes follow the preserved freedoms in Table~\ref{tab:relation-forms}; translations are 20~mm and rotations 0.1~rad.

Table~\ref{tab:e2-observability} reports the maximum finite null response $I_{\max}$, differential null response $N_{\max}$, finite-difference error $E_{\mathrm{FD,max}}$ with step $10^{-5}$, and the numerical rank of the local Jacobian at cutoff $10^{-8}$, compared with the registered $\dim \mathcal{B}_{\gamma}$.

All 64 cases satisfied the $10^{-8}$ null-response and $10^{-6}$ finite-difference thresholds, while all relation-relevant probes remained observable. The observed ranks matched the registered $\dim B_\gamma$ for all eight forms, consistent with the local differential condition in~\eqref{eq:quotient-kernel}. These results support local invariance and differential consistency of the implemented relation maps at the evaluated configurations. The unilateral boundary of the signed-gap/contact form is evaluated separately below.

\subsubsection{Composition and Coordinate-Lock Ablation}
\label{sec:composition-ablation}

A separate algebraic evaluation is performed in a six-dimensional
local displacement space. For each composition, we stack the selected
equality Jacobians and test whether the intersection of their null
spaces matches the kernel of the composed Jacobian; an inequality
$g(x) \geq 0$ is evaluated through its feasible tangent condition
$Dg(x)v \geq 0$.

The three progress compositions in Table~\ref{tab:e2-composition} have
rank/nullity $4/2$, $4/2$, and $3/3$, respectively, with kernel
residuals below $10^{-15}$ across rank tolerances from $10^{-7}$ to
$10^{-9}$. The Plane--contact case has rank 2 and nullity 4, with the
feasible-side probe accepted and the opposite probe rejected.

\begin{table}[t]
\caption{Task-map invariance and differential consistency. $\dagger$: structural zero (coordinate absent from the task map).}
\label{tab:e2-observability}
\centering
\footnotesize
\setlength{\tabcolsep}{3pt}
\renewcommand{\arraystretch}{1.1}
\begin{tabular}{@{}lcccc@{}}
\toprule
Geometric relation form & $I_{\max}$ & $N_{\max}$ & $E_{\mathrm{FD},\max}$ & $\operatorname{rank}(J_\gamma)$ \\
\midrule
Line parallelism & $1.12\mathrm{e}{-16}$ & $0^{\dagger}$ & $1.72\mathrm{e}{-11}$ & 2 \\
\addlinespace[1pt]
Line coincidence & $1.30\mathrm{e}{-16}$ & $1.16\mathrm{e}{-17}$ & $1.11\mathrm{e}{-11}$ & 4 \\
\addlinespace[1pt]
Plane-normal alignment & $1.13\mathrm{e}{-16}$ & $0^{\dagger}$ & $1.76\mathrm{e}{-11}$ & 2 \\
\addlinespace[1pt]
Plane coincidence & $1.26\mathrm{e}{-16}$ & $3.47\mathrm{e}{-17}$ & $1.11\mathrm{e}{-11}$ & 3 \\
\addlinespace[1pt]
Directional alignment & $1.11\mathrm{e}{-16}$ & $0^{\dagger}$ & $1.67\mathrm{e}{-11}$ & 2 \\
\addlinespace[1pt]
Signed gap/contact & $0^{\dagger}$ & $0^{\dagger}$ & $1.00\mathrm{e}{-12}$ & 1 \\
\addlinespace[1pt]
Circle incidence & $8.34\mathrm{e}{-15}$ & $7.24\mathrm{e}{-14}$ & $4.16\mathrm{e}{-10}$ & 2 \\
\addlinespace[1pt]
Circle traversal & $0^{\dagger}$ & $0^{\dagger}$ & $4.16\mathrm{e}{-10}$ & 3 \\
\bottomrule
\end{tabular}
\vspace{-0.1cm}
\end{table}

\begin{table}[t]
\caption{Local 6D rank/nullity and terminal progress errors (mm; median [min, max], five matched seeds).}
\label{tab:e2-composition}
\centering
\footnotesize
\setlength{\tabcolsep}{4pt}
\renewcommand{\arraystretch}{1.1}
\begin{tabular}[c]{@{}lccc@{}}
\toprule
Composition (Lock) & Rank/null & Relation & Lock \\
\midrule
\makecell[l]{Align--insert\\{(mating axis)}} & 4/2 & \makecell{2.91\\{[2.90, 17.68]}} & \makecell{65.67\\{[59.74, 87.68]}} \\
\addlinespace[2pt]
\makecell[l]{Align--traverse\\{(path tangent)}} & 4/2 & \makecell{4.57\\{[4.44, 4.63]}} & \makecell{75.58\\{[75.38, 75.75]}} \\
\addlinespace[2pt]
\makecell[l]{Directional alignment--progress\\{(vertical)}} & 3/3 & \makecell{0.95\\{[0.92, 0.96]}} & \makecell{21.19\\{[21.19, 21.20]}} \\
\bottomrule
\end{tabular}
\vspace{-0.1cm}
\end{table}

\begin{figure*}[!t]
    \centering
    \includegraphics[width=\textwidth]{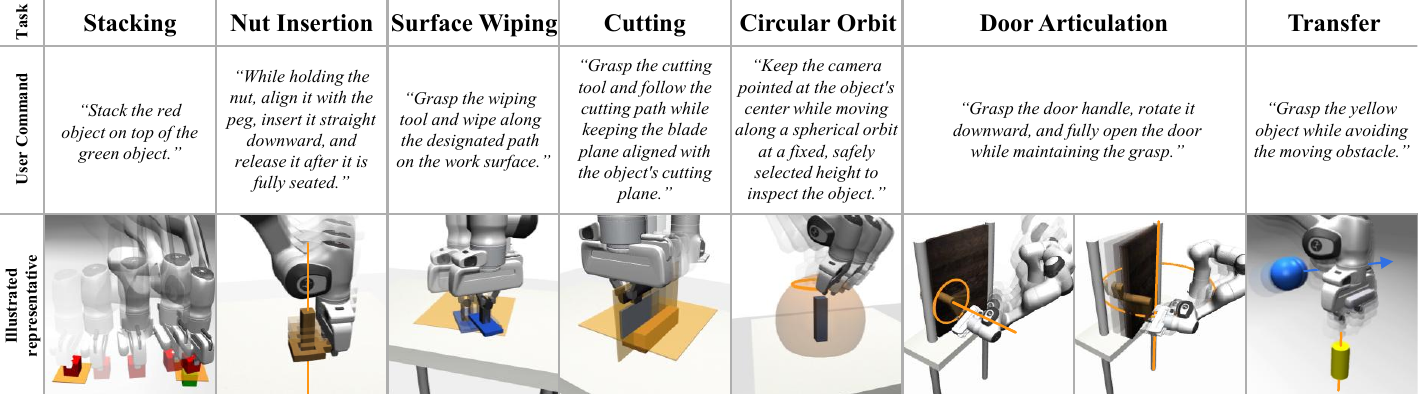}
    \caption{Representative executions of seven tasks with active primitives overlaid. Table~\ref{tab:task-reuse} lists their primitive types and semantic relations. These executions illustrate reuse of registered relation specifications across different entities and primitives.}
    \label{fig:task-reuse}
    \vspace{-0.3cm}
\end{figure*}

Using the abbreviations in Table~\ref{tab:semantic-vocabulary}, Align--insert composes the Line-coincidence alignment of the nut with the seating progress of its contact stage; Align--traverse composes blade-plane coincidence and surface contact with \texttt{TA}; and Directional alignment--progress composes \texttt{MD} with \texttt{CS} in a Stacking transport topology that adds vertical-axis maintenance to the execution in Fig.~\ref{fig:task-reuse}.

In the closed-loop ablations, Table~\ref{tab:e2-composition} tests whether fixing a coordinate preserved by an active relation interferes with a concurrently active progress objective. Terminal progress error is the axial seating error in Align--insert, the path endpoint error in Align--traverse, and the remaining vertical progress in Directional alignment--progress. Relative to the relation condition, the median terminal progress error increased by $62.76$~mm in Nut insertion, $71.00$~mm in Cutting, and $20.24$~mm in Stacking transport.
\begin{table}[!t]
    \caption{Primitives and typed semantic relations in the executions of Fig.~\ref{fig:task-reuse}. Abbreviations are defined in Table~\ref{tab:semantic-vocabulary}.}
    \label{tab:task-reuse}
    \centering
    \footnotesize
    \setlength{\tabcolsep}{4pt}
    \renewcommand{\arraystretch}{1.08}

    \begin{tabularx}{\columnwidth}{
        @{}
        >{\raggedright\arraybackslash}p{0.24\columnwidth}
        >{\raggedright\arraybackslash}p{0.22\columnwidth}
        >{\raggedright\arraybackslash}X
        @{}
    }
    \toprule
    Task & Primitive types & Typed semantic relations \\
    \midrule

    Stacking
    & Plane
    & \texttt{RA}, \texttt{CS}, \texttt{MRP}, \texttt{ALN}, \texttt{EC} \\

    Nut Insertion
    & Line
    & \texttt{RA}, \texttt{CS}, \texttt{MRP}, \texttt{ALN}, \texttt{EC} \\

    Surface Wiping
    & Plane
    & \texttt{RA}, \texttt{CS}, \texttt{MRP}, \texttt{ALN},
      \texttt{EC}, \texttt{MAC}, \texttt{TA} \\

    Cutting
    & Plane
    & \texttt{RA}, \texttt{CS}, \texttt{MRP}, \texttt{ALN},
      \texttt{EC}, \texttt{MAC}, \texttt{TA} \\

    Circular Orbit
    & Circle, Sphere
    & \texttt{ALN}, \texttt{TA} \\

    Door Articulation
    & Line, Circle
    & \texttt{RA}, \texttt{MAC}, \texttt{ACT} \\

    Transfer
    & Point, Line
    & \texttt{RA}, \texttt{CS}, \texttt{MRP}, \texttt{AC} \\

    \bottomrule
    \end{tabularx}

\end{table}
All lock trials were stopped after progress stalled before the objective was completed. The relation condition completed all trials except one Nut trial, which also stalled; its larger terminal error is retained in the reported range. Relation-condition values are the residuals recorded before the completion check; in Stacking transport, all five lie within the $1$~mm completion tolerance.
Under lock, the retained relation errors remained within their evaluation bounds: transverse error was at most $3.30$~mm in Nut insertion, the median maximum blade error was $0.24^{\circ}$ with full contact in Cutting, and direction error was at most $0.74^{\circ}$ in Stacking. The cosine similarity between the configuration-space natural-form forces of the lock and progress policies rounds to $-1.000$ in all three cases, indicating opposing action along the locked coordinate.
These interventions isolate the representation-induced conflict anticipated in Sec. I: fixing a coordinate left free by a relation can obstruct concurrent task progress. This ablation adds a single unnecessary pose constraint; it does not compare against full-pose controllers in general.

\subsubsection{Observed Cross-Task Relation Reuse}

Fig.~\ref{fig:task-reuse} and Table~\ref{tab:task-reuse} illustrate relation reuse across seven task families. \texttt{TA} binds to different tools and paths in Wiping and Cutting, while \texttt{ALN} applies to different primitive pairs in stacking, insertion, orbit, and surface interaction. These examples show observed reuse, but do not establish deterministic task-to-relation mappings, selection stability, unseen composition, or cross-robot generalization.

\subsection{E3: Prompt-Conditioned Relation Selection and Motion}

E3 qualitatively examines how prompt variations change relation selection and executed motion for Nut insertion in the same scene; it does not estimate success rates.

\begin{figure}[t]
    \centering
    \includegraphics[width=\linewidth]{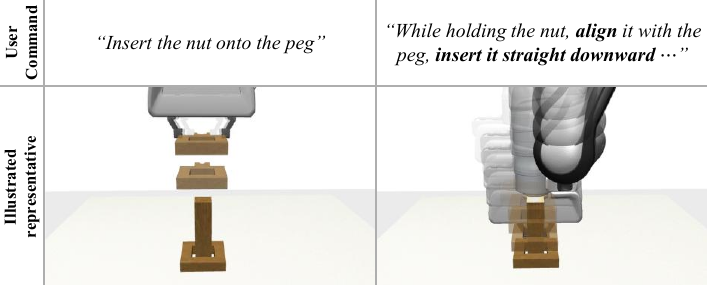}
    \caption{Prompt-conditioned relation selection in Nut insertion. N0 (left) releases after entry alignment; N1 (right) keeps grasp and alignment during descent and releases only after seating.}
    \label{fig:E3}
    \vspace{-0.1cm}
\end{figure}

The N0 prompt, \emph{Insert the nut onto the peg}, does not prescribe a detailed process. N1 requests, \emph{While holding the nut, align it with the peg, insert it straight downward, and release it after it is fully seated}. Both topologies include grasp, support clearance, entry alignment, axis alignment, and key alignment. N1 additionally introduces seat alignment and seated contact during the held-straight insertion phase and places release in a subsequent phase. As a result (Fig.~\ref{fig:E3}), N0 releases after entry alignment, whereas N1 maintains grasp and alignment during descent and releases only once the seated-contact condition is satisfied. For the same entities and primitives, the prompt changed the active relation set and the phase boundary, while the compiler and registry were identical in both cases.

\section{Discussion}
\label{sec:discussion}

The results support encoding a qualitative relation as executable task quantities together with the freedoms it preserves, rather than expanding it into a complete-pose objective. In this view, preserved freedom is part of the relation semantics rather than a degree of freedom left over after controller computation. When each relation excludes irrelevant directions from its task map, composition combines only the distinctions required in the active phase instead of imposing a complete pose objective in advance.

The evaluation has several limitations. It assumes that the world description provides the entities and affordances required to construct the Semantic Topology, so the effects of missing or inaccurate perception and ambiguous object references on topology generation and grounding are not evaluated. Because this work focuses on mapping selected relations and phases to task-visible geometry and control rather than open-ended task decomposition, several representative natural-language instructions explicitly state procedural details. E2 verifies local consistency of the implemented relation maps and their Jacobians for all eight listed geometric relation forms at the evaluated configurations and, as stated in Sec.~\ref{sec:scope}, does not address global feasibility or closed-loop stability. The cross-task reuse and E3 results are representative simulator observations rather than evaluations of repeated sampling or unseen compositions. Although relation specifications are intended to be robot-independent, all current execution experiments use a single simulated Panda; cross-robot rebinding and real-world contact reliability remain unevaluated.

\section{Conclusion}
\label{sec:conclusion}

This work presented a semantic-to-geometric interface that maps language-specified entities, relations, and phases to relation-specific continuous control objectives while preserving relation-unspecified freedoms. A robot-side compiler grounds the validated Semantic Topology in the scene and constructs the corresponding task maps, differentials, and relation-local policies.

The relation-level interface produced fewer critical errors than the geometry-explicit interface (19/60 versus 58/60). Across eight geometric relation forms, the implemented task maps preserved registered null directions and matched their expected differential ranks, while closed-loop ablations showed that unnecessarily fixing a preserved coordinate interferes with concurrent progress. Representative executions further showed reuse of the same relation specifications across different primitives and tasks, and prompt variations changed the active relations and phase boundaries without changing the compiler. These results support compiling relation-visible geometry and preserved freedom together into composable continuous objectives.

\bibliographystyle{IEEEtran}
\bibliography{references}

@article{garrett2021tamp,
  author = {C. R. Garrett and R. Chitnis and R. Holladay and B. Kim and
            T. Silver and L. P. Kaelbling and T. Lozano-P{\'e}rez},
  title = {Integrated Task and Motion Planning},
  journal = {Annual Review of Control, Robotics, and Autonomous Systems},
  volume = {4},
  pages = {265--293},
  year = {2021}
}

@inproceedings{curtis2025proc3s,
  author = {A. Curtis and N. Kumar and J. Cao and T. Lozano-P{\'e}rez and L. P. Kaelbling},
  title = {Trust the {PRoC3S}: Solving Long-Horizon Robotics Problems with {LLMs} and Constraint Satisfaction},
  booktitle = {Proceedings of the 8th Conference on Robot Learning},
  series = {Proceedings of Machine Learning Research},
  volume = {270},
  pages = {1362--1383},
  publisher = {PMLR},
  year = {2025},
  url = {https://proceedings.mlr.press/v270/curtis25a.html}
}

@article{berenson2011tsr,
  author = {D. Berenson and S. S. Srinivasa and J. Kuffner},
  title = {Task Space Regions: A Framework for Pose-Constrained Manipulation Planning},
  journal = {The International Journal of Robotics Research},
  volume = {30},
  number = {12},
  pages = {1435--1460},
  year = {2011}
}

@article{bregier2018pose,
  author = {R. Br{\'e}gier and F. Devernay and L. Leyrit and J. L. Crowley},
  title = {Defining the Pose of Any {3D} Rigid Object and an Associated Distance},
  journal = {International Journal of Computer Vision},
  volume = {126},
  number = {6},
  pages = {571--596},
  year = {2018}
}

@book{dorst2007geometric,
  author    = {L. Dorst and D. Fontijne and S. Mann},
  title     = {Geometric Algebra for Computer Science: An Object-Oriented Approach to Geometry},
  publisher = {Morgan Kaufmann},
  year      = {2007}
}

@article{low2023geometric,
  author  = {T. L{\"o}w and S. Calinon},
  title   = {Geometric Algebra for Optimal Control with Applications in Manipulation Tasks},
  journal = {IEEE Transactions on Robotics},
  volume  = {39},
  number  = {5},
  pages   = {3586--3600},
  year    = {2023}
}

@article{low2025gafro,
  author = {T. L{\"o}w and P. Abbet and S. Calinon},
  title = {{gafro}: Geometric Algebra for Robotics},
  journal = {IEEE Robotics \& Automation Magazine},
  volume = {32},
  number = {3},
  pages = {184--194},
  year = {2025}
}

@article{ratliff2018rmp,
  author = {N. Ratliff and J. Issac and D. Kappler and S. Birchfield and D. Fox},
  title = {Riemannian Motion Policies},
  journal = {arXiv preprint arXiv:1801.02854},
  year = {2018}
}

@article{cheng2021rmpflow,
  author  = {Cheng, Ching-An and Mukadam, Mustafa and Issac, Jan and
             Birchfield, Stan and Fox, Dieter and Boots, Byron and
             Ratliff, Nathan D.},
  title   = {{RMPflow}: A Geometric Framework for Generation of
             Multitask Motion Policies},
  journal = {IEEE Transactions on Automation Science and Engineering},
  year    = {2021},
  volume  = {18},
  number  = {3},
  pages   = {968--987},
  doi     = {10.1109/TASE.2021.3053422}
}

@inproceedings{huang2023voxposer,
  author = {Wenlong Huang and Chen Wang and Ruohan Zhang and Yunzhu Li and
            Jiajun Wu and Li Fei-Fei},
  title = {{VoxPoser}: Composable {3D} Value Maps for Robotic Manipulation with Language Models},
  booktitle = {Proceedings of the 7th Conference on Robot Learning},
  series = {Proceedings of Machine Learning Research},
  volume = {229},
  pages = {540--562},
  publisher = {PMLR},
  year = {2023},
  url = {https://proceedings.mlr.press/v229/huang23b.html}
}

@article{huang2024rekep,
  author = {Wenlong Huang and Chen Wang and Yunzhu Li and Ruohan Zhang and Li Fei-Fei},
  title = {{ReKep}: Spatio-Temporal Reasoning of Relational Keypoint Constraints for Robotic Manipulation},
  journal = {arXiv preprint arXiv:2409.01652},
  year = {2024}
}

@article{tang2025geomanip,
  author = {Weiliang Tang and Jia-Hui Pan and Yun-Hui Liu and Masayoshi Tomizuka and
            Li Erran Li and Chi-Wing Fu and Mingyu Ding},
  title = {{GeoManip}: Geometric Constraints as General Interfaces for Robot Manipulation},
  journal = {arXiv preprint arXiv:2501.09783},
  year = {2025}
}

@article{zhang2026relafford6d,
  author = {Guodong Zhang and Qichen He and Wenyuan Xie and Shaokai Wu and
            Yanbiao Ji and Qiuchang Li and Bayram Bayramli and Yue Ding and Hongtao Lu},
  title = {{RelAfford6D}: Relational {6D} Affordance Graphs for Constraint-Driven Robotic Manipulation},
  journal = {arXiv preprint arXiv:2606.27036},
  year = {2026}
}

@inproceedings{ichter2023saycan,
  author    = {Brian Ichter and Anthony Brohan and Yevgen Chebotar and others},
  title     = {Do As {I} Can, Not As {I} Say: Grounding Language in Robotic Affordances},
  booktitle = {Proceedings of the Conference on Robot Learning},
  series    = {Proceedings of Machine Learning Research},
  volume    = {205},
  pages     = {287--318},
  year      = {2023}
}

@inproceedings{singh2023progprompt,
  author    = {Ishika Singh and Valts Blukis and Arsalan Mousavian and others},
  title     = {{ProgPrompt}: Generating Situated Robot Task Plans Using Large Language Models},
  booktitle = {Proceedings of the IEEE International Conference on Robotics and Automation},
  year      = {2023},
  doi       = {10.1109/ICRA48891.2023.10161317}
}

@inproceedings{liang2023code,
  author    = {Jacky Liang and Wenlong Huang and Fei Xia and others},
  title     = {Code as Policies: Language Model Programs for Embodied Control},
  booktitle = {Proceedings of the IEEE International Conference on Robotics and Automation},
  year      = {2023},
  doi       = {10.1109/ICRA48891.2023.10160591}
}

@inproceedings{driess2023palme,
  author    = {Danny Driess and Fei Xia and Mehdi S. M. Sajjadi and others},
  title     = {{PaLM-E}: An Embodied Multimodal Language Model},
  booktitle = {Proceedings of the International Conference on Machine Learning},
  series    = {Proceedings of Machine Learning Research},
  volume    = {202},
  year      = {2023}
}

@inproceedings{berenson2009constraint,
  author    = {Dmitry Berenson and Siddhartha S. Srinivasa and Dave Ferguson and James Kuffner},
  title     = {Manipulation Planning on Constraint Manifolds},
  booktitle = {Proceedings of the IEEE International Conference on Robotics and Automation},
  pages     = {625--632},
  year      = {2009}
}

@article{englert2018taskmanifold,
  author  = {Peter Englert and Isabel M. P. G. Rolf and Marc Toussaint},
  title   = {Learning Task Manifolds for Constrained Object Manipulation},
  journal = {Autonomous Robots},
  volume  = {42},
  number  = {1},
  pages   = {159--174},
  year    = {2018},
  doi     = {10.1007/s10514-017-9643-z}
}

@inproceedings{toussaint2015lgp,
  author    = {Marc Toussaint},
  title     = {Logic-Geometric Programming: An Optimization-Based Approach to Combined Task and Motion Planning},
  booktitle = {Proceedings of the International Joint Conference on Artificial Intelligence},
  year      = {2015}
}

@inproceedings{toussaint2022socmpc,
  author    = {Marc Toussaint and Jason Harris and Jung-Su Ha and Danny Driess and Wolfgang H{\"o}nig},
  title     = {Sequence-of-Constraints {MPC}: Reactive Timing-Optimal Control of Sequential Manipulation},
  booktitle = {Proceedings of the IEEE/RSJ International Conference on Intelligent Robots and Systems},
  pages     = {13753--13760},
  year      = {2022}
}

@article{vanwyk2022fabrics,
  author  = {Karl {Van Wyk} and Mandy Xie and Anqi Li and Muhammad Rana and Buck Babich and Bryan Peele and Qian Wan and Ireti Akinola and Balakumar Sundaralingam and Dieter Fox and Byron Boots and Nathan Ratliff},
  title   = {Geometric Fabrics: Generalizing Classical Mechanics to Capture the Physics of Behavior},
  journal = {IEEE Robotics and Automation Letters},
  volume  = {7},
  number  = {2},
  pages   = {3202--3209},
  year    = {2022}
}

@misc{rosmcp,
  author       = {{RobotMCP}},
  title        = {{ros-mcp-server}: Connect {AI} Models with Robots Using {MCP} and {ROS}},
  year         = {2025},
  howpublished = {\url{https://github.com/robotmcp/ros-mcp-server}},
  note         = {Accessed: Mar. 2026}
}

\end{document}